\documentclass[runningheads,a4paper]{llncs}

\usepackage{amssymb}
\usepackage{graphicx}
\usepackage{amsmath}
\usepackage{times}
\usepackage{url}
\usepackage{color}

\usepackage{tabularx,ragged2e,booktabs,caption}

\DeclareMathOperator*{\argmax}{arg\,max}

\graphicspath{{./fig/}}

\begin{document}

\mainmatter  

\title{Deep Projective 3D Semantic Segmentation}

\titlerunning{Deep Projective 3D Semantic Segmentation}

%
%
\author{Felix J\"aremo Lawin, Martin Danelljan, Patrik Tosteberg, \\ Goutam Bhat, Fahad Shahbaz Khan, Michael Felsberg}
\authorrunning{Felix J\"aremo Lawin et al.}
\institute{Computer Vision Lab, Dept.\ of Electrical Engineering, Link\"oping University}

%
%

\maketitle

\begin{abstract}
Semantic segmentation of 3D point clouds is a challenging problem with numerous real-world applications. While deep learning has revolutionized the field of image semantic segmentation, its impact on point cloud data has been limited so far. Recent attempts, based on 3D deep learning approaches (3D-CNNs), have achieved below-expected results. Such methods require voxelizations of the underlying point cloud data, leading to decreased spatial resolution and increased memory consumption. Additionally, 3D-CNNs greatly suffer from the limited availability of annotated datasets.

In this paper, we propose an alternative framework that avoids the limitations of 3D-CNNs. Instead of directly solving the problem in 3D, we first project the point cloud onto a set of synthetic 2D-images. These images are then used as input to a 2D-CNN, designed for semantic segmentation. Finally, the obtained prediction scores are re-projected to the point cloud to obtain the segmentation results. We further investigate the impact of multiple modalities, such as color, depth and surface normals, in a multi-stream network architecture. Experiments are performed on the recent Semantic3D dataset. Our approach sets a new state-of-the-art by achieving a relative gain of $7.9 \%$, compared to the previous best approach.
\end{abstract}

\section{Introduction}


The rapid development of 3D acquisition sensors, such as LIDARs and RGB-D cameras, has lead to an increased demand for automatic analysis of 3D point clouds. In particular, the ability to automatically categorize each point into a set of semantic labels, known as semantic point cloud segmentation, has numerous applications such as scene understanding and robotics. While the problem of semantic segmentation of 2D-images has gained a considerable amount of attention in recent years, semantic segmentation of point clouds has received little interest despite its significance. In this paper, we propose a framework for semantic segmentation of point clouds that greatly benefits from the recent developments in semantic image segmentation. 

With the advent of deep learning, many tasks within computer vision have seen a rapid progress, including semantic segmentation of images. The key factors for this development are the introductions of large labeled datasets \cite{imagenet_cvpr09} and GPU implementations of Convolutional Neural Networks (CNNs). However, CNNs have not yet been successfully applied for semantic segmentation of 3D point clouds due to several challenges. In contrast to the regular grid-structure of image data, point clouds are in general sparse and unstructured. A common strategy is to resort to voxelization in order to directly apply CNNs in 3D. This introduces a radical increase in memory consumption and leads to a decrease in resolution. Additionally, labeled 3D data, which is crucial for training CNNs, is scarce due to difficulties in data annotation.

In this work, we investigate an alternative approach that avoids the aforementioned difficulties induced by 3D CNNs. As our first contribution, we propose a framework for 3D semantic segmentation that exploits the advantages of deep image segmentation approaches. The point cloud is first projected onto a set of synthetic images, which are then used as input to the deep network. The resulting pixel-wise segmentation scores are re-projected into the point cloud. The semantic label for each point is then obtained by fusing scores over the different views. As our second contribution, we investigate the impact of different input modalities, such as color, depth and surface normals, extracted from the point cloud. These modalities are fused in a multi-stream network architecture to obtain the final prediction scores. 

Compared to semantic segmentation methods based on 3D CNNs \cite{maturana2015voxnet}, our approach has two major advantages. Firstly, our method benefits from the abundance of the already existing data sets for image segmentation and classification, such as ImageNet \cite{imagenet_cvpr09} and ADE20K \cite{zhou2017scene}. This significantly reduces, or even eliminates the need of 3D data for training purposes. Secondly, by avoiding the large memory complexity induced by voxelization, our method achieves a higher spatial resolution which enables better segmentation quality. 

We perform qualitative and quantitative experiments on the recently introduced Semantic3D dataset \cite{hackel2017semantic3d}. We show that different modalities contain complementary information and their fusion significantly improves the final segmentation performance. Further, our approach sets a new state-of-the-art performance on the Semantic3D dataset, outperforming both classical machine learning methods and 3D-CNN based approaches. Figure~\ref{fig:seg} shows an example segmentation result using our method.

\section{Related Work}

The task of semantic point cloud segmentation has received an increasing amount of attention due to the rapid development of sensors capable of capturing high-quality 3D data. RGB-D cameras, such as the Microsoft Kinect, have become popular for robotics and computer vision tasks. While RGB-D cameras are more suitable for indoors environments, terrestrial  laser scanners capture large-scale point clouds for both indoors and outdoors applications. Both RGB-D cameras and modern laser scanners are capable of capturing color in association with the 3D information using calibrated RGB cameras. Besides visualization, this additional information is highly useful for automated analysis and processing of point clouds. While color is not a necessity for our approach, it alleviates the task of semantic segmentation and enables the use of large-scale image datasets.

Most previous works \cite{AnguelovCVPR05,HackelISPRS16,KahlerICCV13,MartinovicICCV15,KimICCV13} in 3D semantic segmentation apply a combination of (i) hand-crafted features, (ii) discriminative classifiers and (iii) spatial smoothness models. In this setting, the construction of discriminative 3D-features (i) is arguably the most important task. Popular alternatives include features based on the 3D structure tensor \cite{HackelISPRS16,WolfICRA15,KahlerICCV13,AnguelovCVPR05}, histogram-based descriptors \cite{HackelISPRS16,MartinovicICCV15,KahlerICCV13} such as Spin Images \cite{JohnsonPAMI99} and SHOT \cite{SaltiCVIU14}, and simple color features \cite{WolfICRA15,MartinovicICCV15,KahlerICCV13}. The classifiers (ii) are often based on maximum margin methods \cite{KimICCV13,AnguelovCVPR05} or employ random forests \cite{HackelISPRS16,KahlerICCV13,MartinovicICCV15}. To utilize spatial correlation between semantic labels (iii), many methods apply graphical models, such as the Conditional Random Field (CRF) \cite{WolfICRA15,AnguelovCVPR05,KimICCV13}.

Recently, deep convolutional neural networks (CNNs) have been successfully applied for semantic segmentation of 2D images \cite{long2015fully}. Their main strength is the ability to learn high-level discriminative features, which eliminates the need of hand-designed representations. The rapid progress of deep CNNs for a variety of computer vision problems is generally attributed to the introduction of large-scale datasets, such as ImageNet \cite{imagenet_cvpr09}, and improved performance for GPU computing.

Despite its success for image data, the application of CNNs to 3D point cloud data \cite{HuangICPR16,QiCVPR16,WuCVPR15} have been severely hindered due to several important factors. Firstly, a point cloud does not have the neighborhood structure of an image. The data is instead sparse and scattered. As a consequence, CNN-based methods resort to voxelization strategies of the underlying point cloud data to enable 3D-convolutions to be performed (3D-CNNs). Secondly, voxelization have several disadvantages, including loss of spatial resolution and large memory requirements. 3D-CNNs are therefore restricted to small volumetric models or processing data in many smaller chunks, which limits the use of context. Thirdly, annotated 3D data is extremely limited, especially for the 3D semantic segmentation task. This greatly limits the power of CNNs for semantic segmentation of generic 3D point clouds. 

In contrast, our approach avoids these short comings by projecting the point cloud into dense 2D image representations, thus removing the need for voxelizations. The 2D images can then be efficiently processed using 2D convolutions. Also, performing segmentation in image space allows us to leverage well developed 2D segmentation techniques as well as large amount of annotated data.

\section{Method}

\begin{figure}[!t]
\begin{center}
  \includegraphics[width=0.9\columnwidth]{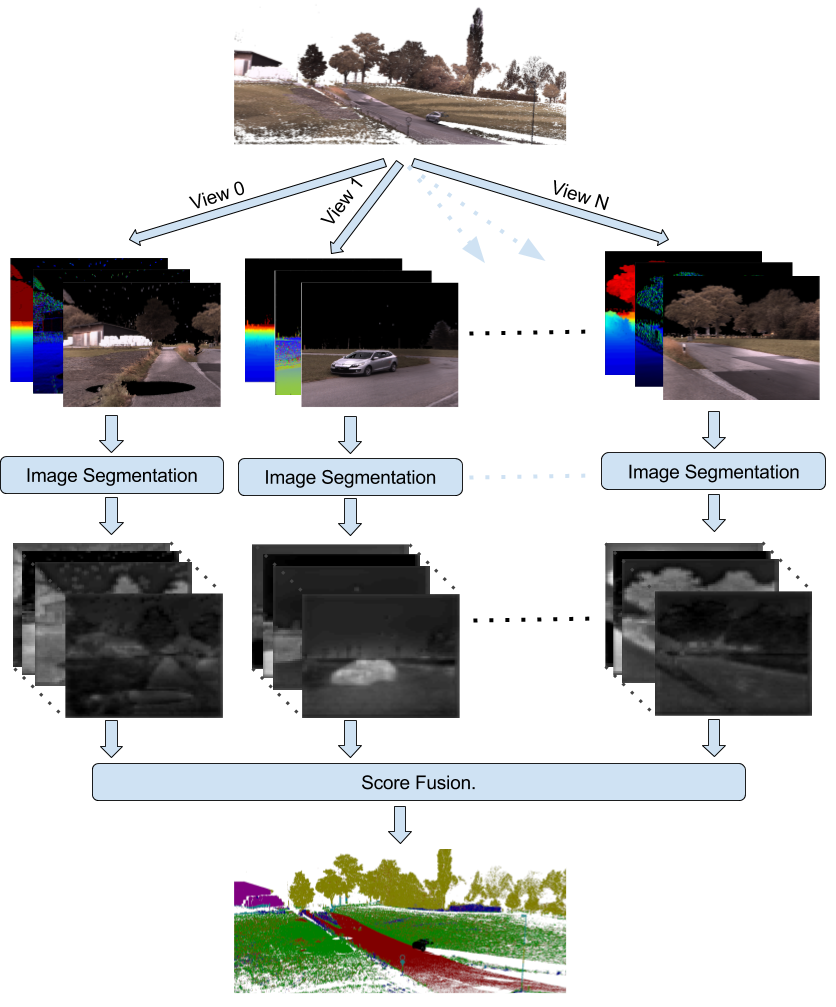}
  \end{center}
  \caption{An overview of the proposed method. The input point cloud is projected into multiple virtual camera views, generating 2D color, depth and surface normal images. The images for each view are processed by a multi-stream CNN for semantic segmentation. The output prediction scores from all views are fused into a single prediction for each point, resulting in a 3D semantic segmentation of the point cloud.}
  \label{fig:overview}
\end{figure}

In this section we present our method for point cloud segmentation. The input is an unstructured point cloud and the objective is to assign a semantic label to each point. In our method we render the point cloud from different views by projecting the points into synthetic images. We render color, depth and other attributes extracted from the point cloud. The images are then processed by a CNN for image-based semantic segmentation, providing a prediction scores for the predefined classes in every pixel. We make the final class selection from the aggregated prediction scores, using all images where the particular points are visible. An overview of the method is illustrated in Figure~\ref{fig:overview}. A more detailed description is provided in the following sections.

\subsection{Render views}
\label{sec:render}

\begin{figure}[!t]
    \begin{center}
        \includegraphics[width=0.49\columnwidth]{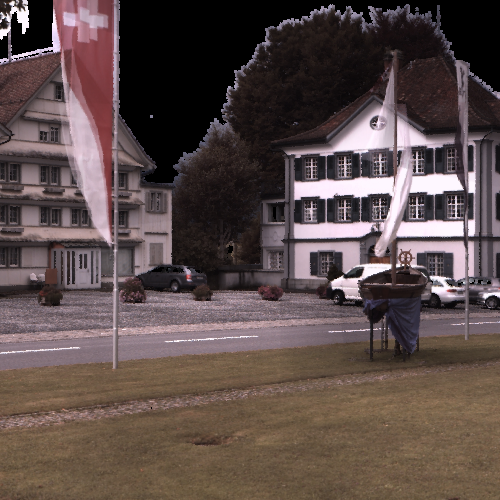}
        \includegraphics[width=0.49\columnwidth]{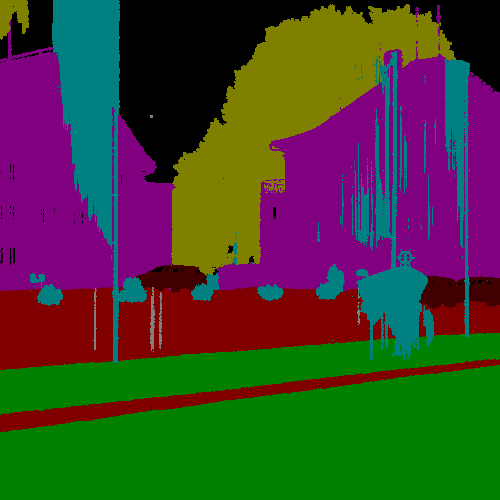}
    \end{center}
    \caption{Example of rendering output. Left: color image. Right: label image.}
    \label{fig:render}
\end{figure}

The objective of the point cloud rendering is to produce structured 2D-images that are used as input to a CNN-based semantic segmentation algorithm. A variety of information stemming from the point cloud can be projected onto the synthetic images. In this work we particularly investigate the use of depth, color, and normals. However, the approach can be trivially extended to other features such as HHA \cite{gupta2014learning} and other local information extracted from the point cloud. In order to map the semantic information back to the 3D points, we also need to keep track of the visibility of the projected points. 

Our choice of rendering technique is a variant of point splatting \cite{szeliski10,zwicker2001surface}, where the points are projected with a spread function into the image plane. While other rendering techniques, such as surface reconstruction as in \cite{kazhdan2013screened}, require demanding preprocessing steps of the point cloud in 3D space, splatting could be completely processed in image space. This further enables efficient and easily parallelizable implementations, which is essential for large-scale or dense point clouds.
 




Splatting-based rendering is performed by first projecting each 3D-point ${\bf x}_i$ of the point cloud into the image coordinates ${\bf y}_i$ of a virtual camera. The projected points are stored along with their corresponding depth values $z_i$ and feature vectors ${\bf c}_i$. The latter can include, e.g.,\ the RGB-color and normal vector of the point ${\bf x}_i$. The projection of a 3D-point is distributed by a Gaussian point spread function in the image plane,
\begin{equation}
	w_{i,j} = G(y_i - p_j, \sigma^2) \,.
	\label{eq:splatt_weight}
\end{equation}
Here, $w_{i,j}$ is the contributed weight of point $x_i$ to pixel $j$ in the projected image. It is obtained by evaluating an isotropic Gaussian kernel $G$ with scale $\sigma^2$ at the pixel location $p_j$. In order to reduce computational complexity, the kernel is truncated at a distance $r$. However, point spread functions, which originate from different surfaces, may still intersect in the image plane. Thus, the visibility of the projected points needs to be determined to avoid contributions of occluded surfaces.  Moreover, the sensor data may contain significant foreground noise, such as scanning artifacts, which complicates this task. The challenge is to exclude the contribution from the noise and the occluded surfaces in the rendering process. 



In traditional splatting \cite{zwicker2001surface}, the resulting pixel value is obtained from the weighted average of the point spread functions in an accumulated fashion, using the weights $w_{i,j}$. If the depth of a new point significantly differs from the current weighted average, the pixel depth is either re-initialized with the new value if the point is closer than a specific threshold, or discarded if it is further away \cite{zwicker2001surface}. However, this implies that the resulting pixel value depends on both the threshold value and the order in which the points are projected. Furthermore, noise in the foreground will have significant impact on the resulting images, as it is always rendered.

Similar to the method proposed in \cite{ogniewski17}, we perform mean-shift clustering \cite{szeliski10} of the projected points in each pixel with respect to the depth $z_i$ weighted with $w_{i,j}$ using a Gaussian kernel density estimator $G(d, s^2)$, where $s^2$ denotes the kernel width. Starting from the depth value $d^0_i = z_i$ for each point $i \in I_j$ that contributes to the current pixel $j$, $I_j = \{i: \|p_j - y_i \| < r \}$, the following expression is iterated until convergence

\begin{equation}
	d^{n+1}_i = \frac{\sum_{i \in I_j} w_{i,j} G(d^n_i - z_i,s^2) z_i}{\sum_{i \in I_j} w_{i,j}G(d^n - z_i,s^2) } \,.
	\label{eq:meanshift}
\end{equation}

The iterative process determines a set of unique cluster centers $\{d_k\}^K_1$ from the converged iterates $\{d_i^N\}_{i \in I_j}$. The kernel density of cluster center $d_k$ is given by,

\begin{equation}
	v_k = \frac{\sum_{i \in I_j} w_{i,j} G(d_k - z_i,s^2)}{\sum_{i \in I_j} w_{i,j}} \,.
	\label{eq:kde}
\end{equation}

We rank the clusters with respect to the kernel density estimates and the cluster centers,
\begin{equation}
	s_k = v_k + \frac{D}{d_k} \,.
	\label{eq:cluster_rank}
\end{equation}
Here, the weight $D$ rewards clusters that are near the camera. It is set such that foreground noise and occluded points are not rendered. We chose the optimal cluster as $\tilde{k} = \argmax_k s_k$ and set the depth value of pixel $j$ to the corresponding cluster center $d_{\tilde{k}}$. The feature value is calculated as the weighted average, where the weight is determined by the proximity to the chosen cluster,
\begin{equation}
{\bf c}_{\tilde{k}} = \frac{\sum_{i \in I_j} w_{i,j} G(d_{\tilde{k}} - z_i,s^2) {\bf c}_i}{\sum_{i \in I_j} w_{i,j}G(d_{\tilde{k}} - z_i,s^2) }\,.
\label{eq:feature}
\end{equation}
Since the indices $i \in I_j$ of the contributing points $i$ are stored, it is trivial to map the semantic segmentation scores produced by the CNN back to the point cloud itself. 

An example of the rendering output is shown in Figure~\ref{fig:render}.

\subsection{Deep Multi Stream Image Segmentation}
\begin{figure}[!t]
\begin{center}
  \includegraphics[width=0.99\columnwidth] {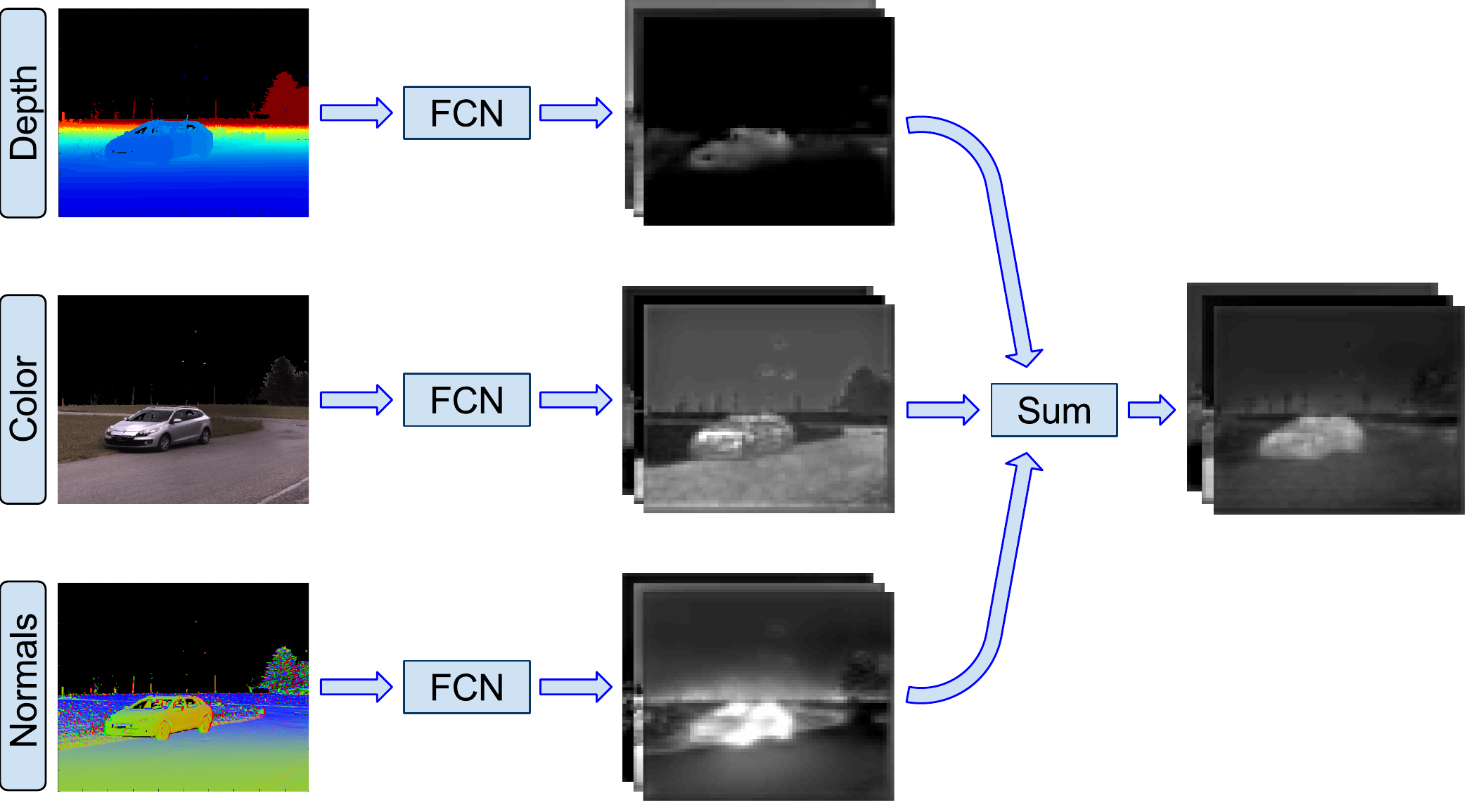}
  \end{center}
  \caption{Illustration of the proposed multi-stream architecture for 2D semantic segmentation. Each input stream is processed by a Fully Convolutional Network\cite{long2015fully}. The prediction scores from each stream are summed to get the final prediction.}
  \label{fig:ms_net}
\end{figure}
Following the current success of deep learning algorithms we deploy a CNN-based algorithm for performing semantic segmentation on the rendered images. We consider using multiple input modalities, which are combined using a multi-stream architecture \cite{DBLP:conf/nips/SimonyanZ14}. The predictions from the streams are fused in a sum layer, as proposed in \cite{DBLP:conf/cvpr/FeichtenhoferPZ16}. The full multi stream network can thus be trained end-to-end. However, note that our pipeline is agnostic to the applied image semantic segmentation approach. 

In our method, each stream is processed using a Fully Convolutional Network (FCN) \cite{long2015fully}. However, as previously mentioned, any CNN architecture can be employed. 
The FCN is based on the popular VGG16 network \cite{Simonyan14c}. The weights in each stream are initialized by pre-training on the ImageNet dataset \cite{imagenet_cvpr09}. In this work, we investigate different combinations of input streams, namely color, depth, and surface normals. While the RGB-stream naturally benefits from pre-training on ImageNet, this is also the case for the depth stream. Previous work \cite{eitel2015multimodal} has shown that a 3-channel jet colormap representation of the depth image better benefits from pre-training on RGB datasets, such as ImageNet. Finally, we also consider surface normals as input to a separate network stream. For this purpose, we deploy an efficient algorithm for approximate normals computation, which is based on direct differentiation of the depth map.

\subsection{Score fusion}
The deep network outputs a prediction score for each class for every pixel in the image. The scores from each rendered view are mapped to the corresponding 3D points using the indices $i \in I_j$ as described in section \ref{sec:render}. We fuse the scores by computing the sum over all projections. Finally, the points are assigned the labels corresponding to the largest sum of scores.


\section{Experiments}

\subsection{Dataset}
We conduct our experiments on the dataset Semantic3D \cite{hackel2017semantic3d}, which provides a set of large scale 3D point clouds of outdoor environments. The point clouds were acquired by a laser scanner and include both urban and rural scenes. Colorization was performed using a cube map generated from a set of high-resolution camera images. In total, the dataset contains 30 separate scans and over 4 billion 3D-points. The points are labeled with 8 different semantic classes: man-made terrain, natural terrain, high vegetation, low vegetation, buildings, hard scape, scanning artifacts, and cars.

\subsection{Experimental setup}

\subsubsection{View selection}
In order to fully cover the point clouds in the rendered views, we collect images by rotating the camera $360^\circ$ around a fix vertical axes. For each $360^\circ$ rotation, we use 30 camera views at equally spaced angles. For each point cloud, we generate four such scans with different pitch angles and translations of the camera, resulting in a total of 120 camera views. To maintain a certain amount of contextual information, we remove images where where more than $10\%$ of the pixels have a depth less than five meters. Furthermore, images with less than $5\%$ coverage were discarded.

\subsubsection{Network setup and training}
For the training we generated ground truth label images by selecting the most commonly occurring label in the optimal cluster from section \ref{sec:render}. An example is shown in Figure~\ref{fig:render}. In addition to the 8 provided classes, we also included a 9th background class to label empty pixels, i.e pixels without any intersecting point spread functions. We generated training data from the training set provided by Semantic3D \cite{hackel2017semantic3d}, consisting of 15 point clouds from different scenes. Our training data set consists of 3132 labeled images including color, jet visualization of the depth, and surface normals. 

We investigate the proposed multi stream approach using color, depth and surface normals streams as input. In order to determine the contribution of each input stream we also evaluate network configurations with a single stream. Since some point clouds may not have color information we also investigate a multi stream approach without the color stream. All network configurations are listed in table \ref{tab:networks}.

\begin{table}
	\centering
	     \caption{Network configurations with input streams in the left column} \label{tab:networks}     
\begin{tabular}{l@{~}|c@{~~}c@{~~}c@{~~}c@{~~}c}  
\toprule       & {\bf RGB} & {\bf D} & {\bf N} & {\bf RGB+D+N} & {\bf D+N}  \\ \midrule  
Color	& X &  &  & X & \\ 
Depth jet &	& X & & X & X \\    
Surface normals &  & & X & X & X \\ \bottomrule  
     \end{tabular}
\end{table}

All network configurations were trained using the same training parameters. We trained for 45 epochs with a batch size of 16. The initial learning rate was set to $0.0001$ and divided by two every tenth epoch. Following the recommendations from \cite{liu2015parsenet}, we used a momentum of 0.99. The networks were trained using MatConvNet \cite{vedaldi15matconvnet}.


\begin{figure}[!t]
    \begin{center}
        \includegraphics[width=0.49\columnwidth] {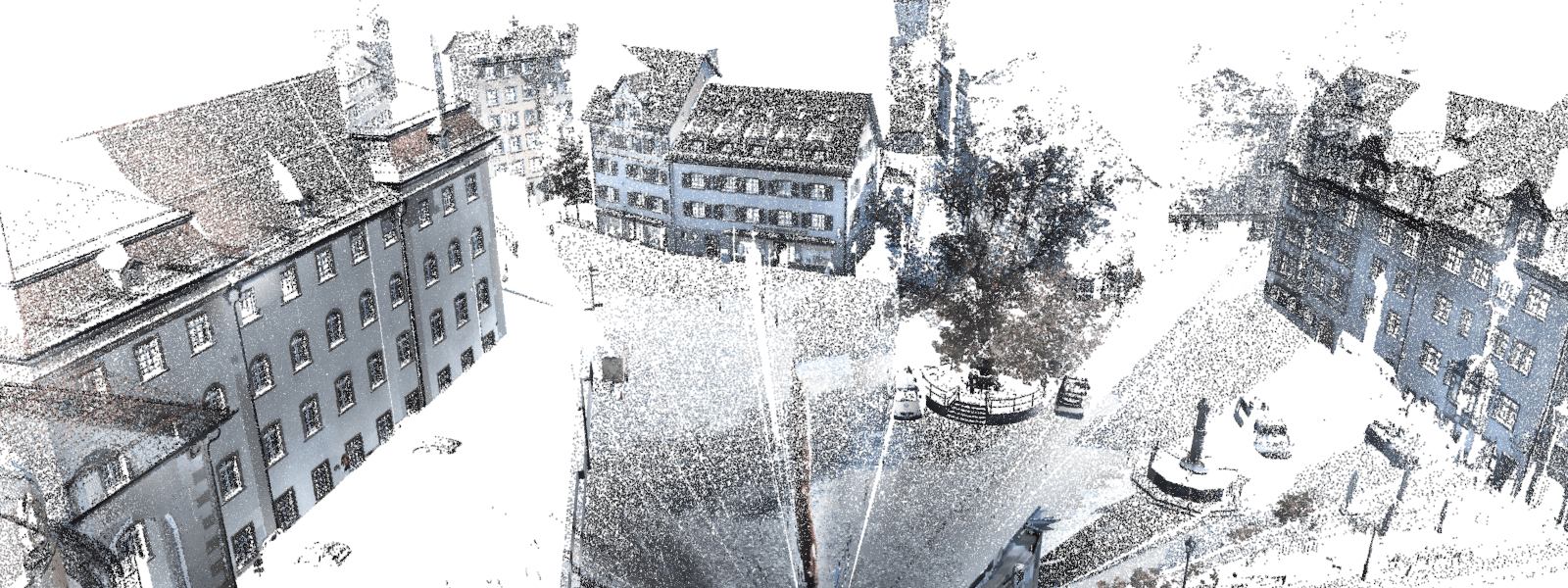}
        \includegraphics[width=0.49\columnwidth] {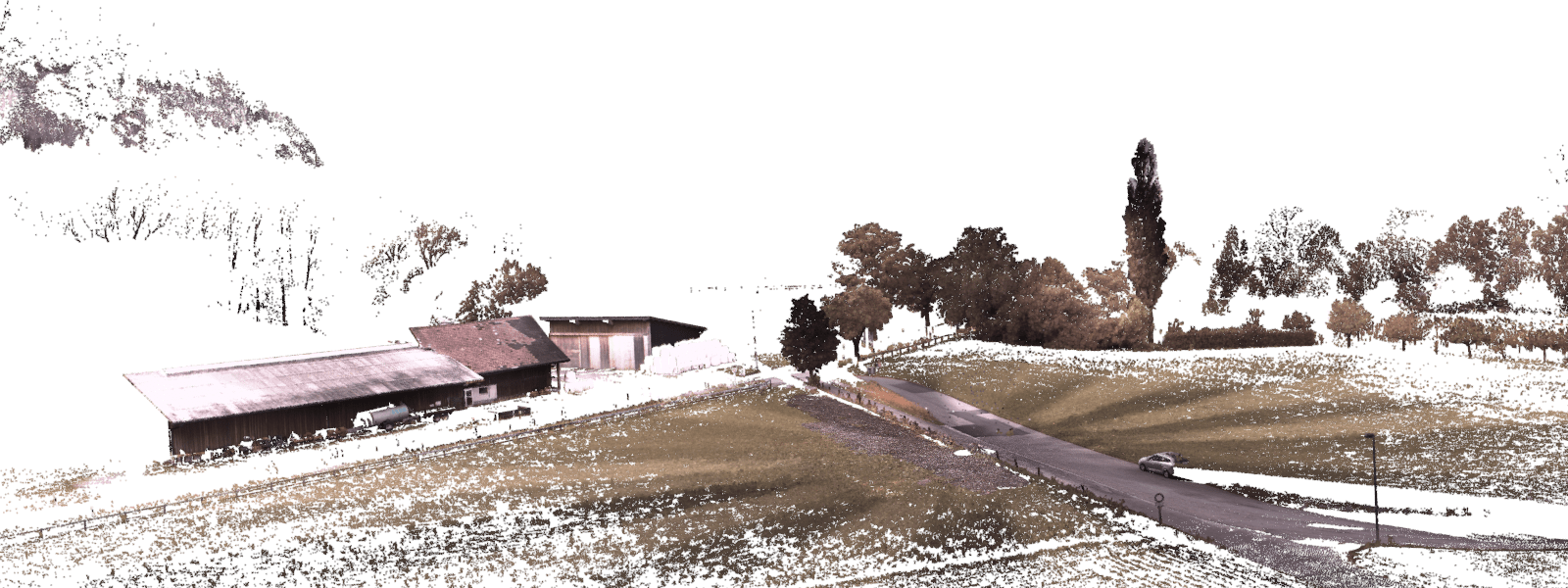}
        \includegraphics[width=0.49\columnwidth] {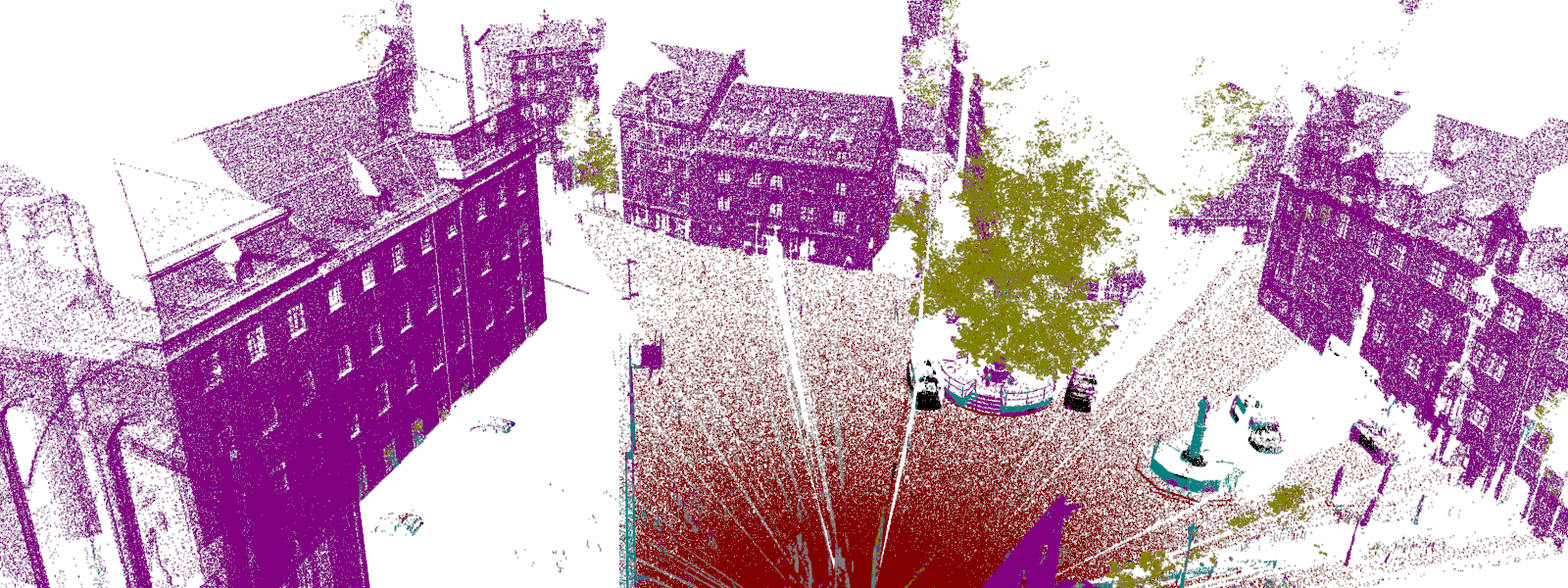}
        \includegraphics[width=0.49\columnwidth] {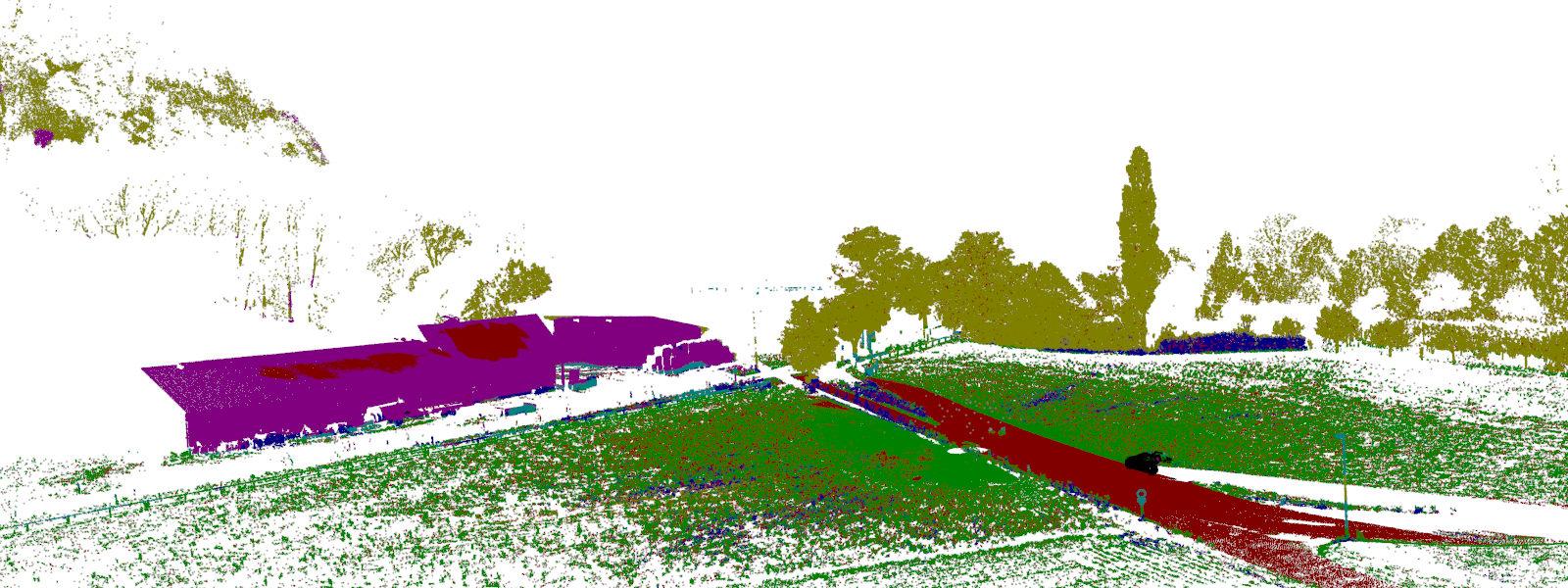}
    \end{center}
    \caption{Qualitative results. Top: input point clouds. Bottom: Segmentation output using our proposed {\bf RGB+D+N} network.}
    \label{fig:seg}
\end{figure}

\subsection{Results and Discussions}
We evaluated our method for the different network configurations on the reduced test set provided by Semantic3D. The test set consists of four point clouds, containing ~80 million points in total. All points are assigned a class label $j$, which is compared to the ground truth label $i$. A confusion matrix $C$ is constructed, were each entry $c_{ij}$ denotes the number of points with the ground truth label $i$ that are assigned the label $j$. The quantitative measure provided by the benchmark \cite{hackel2017semantic3d} is the intersection over union for each class $i$, given by 
\begin{equation}
\text{IoU}_i = \frac{c_{ii}}{c_{ii}+\sum_{j \neq i} c_{ij} + \sum_{k \neq i} c_{kj}}\,.
\end{equation}
The over all accuracy is also provided and is given by
\begin{equation}
\text{IoU} = \frac{\sum_{i} c_{ii}}{\sum_{j}\sum_{jk} c_{jk}}\,.
\end{equation}

The evaluation results are shown in table \ref{tab:reduced}. The single-stream network with RGB and surface normals as input performs significantly better than the single-stream depth network. However, the three streams seem to provide complementary information, and give a significant gain in performance when used together. Our best multi-stream approach significantly improves over the previous state-of-the art method \cite{hackel2016fast}. Also our multi-stream approach without the color stream obtains results comparable to the previous state-of-the, showing that our method is applicable even if color information is absent. Interestingly, even our single-stream approaches with only RGB or surface normals as input achieves a remarkable gain compared to the 3D-CNN based VoxNet \cite{hackel2017semantic3d}. Figure~\ref{fig:seg} shows some qualitative results on the test set using our multi-stream {\bf RBG+D+N} network. 

Note that we are using a simple heuristic for generating camera views, and a basic segmentation network trained on limited data. Yet, we obtain very promising results. Replacing these blocks with better alternatives should improve the results even further. However, this is outside the scope of this paper.

\begin{table}
	\centering
	     \caption{Benchmark results on the reduced test set in Semantic3D  \cite{hackel2017semantic3d}. IoU for categories (1) man-made terrain, (2) natural terrain, (3) high vegetation, (4)
             low vegetation, (5) buildings, (6) hard scape, (7) scanning artefacts, (8) cars.} \label{tab:reduced}     
\begin{tabular}{l@{~}c@{~~}c@{~~}c@{~~}c@{~~}c@{~~}c@{~~}c@{~~}c@{~~}c@{~~}c}  
\toprule       & {\bf Avg IoU} & {\bf OA} & {\bf IoU1} & {\bf IoU2} & {\bf IoU3} & {\bf IoU4} & {\bf IoU5} & {\bf IoU6} & {\bf IoU7} & {\bf IoU8}  \\ \midrule  
TML-PCR\cite{montoya2014mind}	& 0.384	& 0.740	& 0.726	& 0.730	& 0.485	& 0.224	& 0.707	& 0.050	& 0.000	& 0.150  \\ 
DeepNet\cite{hackel2017semantic3d} &	0.437	& 0.772	& 0.838	& 0.385	& 0.548	& 0.085	& 0.841	& 0.151	& 0.223	& 0.423 \\    
TLMC-MSR\cite{hackel2016fast} & 0.542 & 0.862	& {\bf 0.898} &	0.745 &	0.537	& 0.268 &	0.888 &	0.189 &	{\bf 0.364} &	0.447 \\\midrule   {\bf Ours RGB} & 0.515 & 0.854 & 0.759 & 0.791 & 0.720 & {\bf 0.335} & 0.857 & 0.209 & 0.123 & 0.326 \\
{\bf Ours D} & 0.262 &	0.662	& 0.281	& 0.468	& 0.395	& 0.179	& 0.763	& 0.006	& 0.001	& 0.000 \\
{\bf Ours N} & 0.511 & 0.846 & 0.815 & 0.622 & 0.679 & 0.164 &	0.903 & {\bf 0.251} & 0.186 & 0.470 \\
         {\bf Ours RGB+D+N} & {\bf 0.585} & {\bf 0.889} & 0.856 &	{\bf 0.832}	& {\bf 0.742}	& 0.324	& 0.897	& 0.185	& 0.251	& {\bf 0.592} \\
         {\bf Ours D+N} & 0.543	& 0.872	& 0.839	& 0.736	& 0.717	& 0.210	& {\bf 0.909}	& 0.153	& 0.204	& 0.574 \\
         \bottomrule
     \end{tabular}
\end{table}


\section{Conclusion}

We propose an approach for semantic segmentation of 3D point clouds that avoids the limitations of 3D-CNNs. Our approach first projects the point cloud onto a set of synthetic 2D-images. The corresponding images are then used as input to a 2D-CNN for semantic segmentation. Consequently, the segmentation results are obtained by re-projecting the prediction scores to the point cloud. We further investigate the impact of multiple modalities in a multi-stream deep network architecture. Experiments are performed on the Semantic3D dataset. Our approach outperforms existing methods and sets a new state-of-the-art on this dataset.

\bibliographystyle{splncs03}
\bibliography{local}

\end{document}